\documentclass{research}

\title{Reducing the Safety Tax in LLM Safety Alignment with On-Policy Self-Distillation}

\author[1]{Yu Fu}
\author[1]{Longxuan Yu}
\author[1]{Haz Sameen Shahgir}
\author[2]{Zhipeng Wei}
\author[3, *]{Hui Liu}
\author[2,4]{N. Benjamin Erichson}
\author[1]{Yue Dong}

\affiliation[1]{UC Riverside}
\affiliation[2]{International Computer Science Institute}
\affiliation[3]{Microsoft}
\affiliation[4]{Berkeley Lab}

\correspondence{Yue Dong @ \email{yue.dong@ucr.edu}}

\metadata[Code]{\url{https://github.com/FYYFU/OPSA}}

\abstract{Safety alignment often improves robustness to harmful queries at the cost of reasoning ability, a tradeoff known as the \emph{safety tax}. A common cause is distributional mismatch: supervised fine-tuning trains the target model on safety demonstrations produced by humans, external models, or fixed self-generated traces, rather than on trajectories sampled from its own policy. We identify off-policy training mismatch as a second source of this tax and study on-policy self-distillation for safety alignment, which we call OPSA. The model generates its own rollouts and receives dense per-token KL supervision from a frozen teacher copy of itself conditioned on a privileged safety context. Because this teacher must be safer than the sampled student trajectory, we introduce \emph{teacher flip rate}: a criterion that measures how often a privileged context converts unsafe responses into safe ones. We use this signal to search for contexts that activate latent safety reasoning rather than merely elicit safe-looking demonstrations. Across two reasoning-model families and five model scales, OPSA achieves a stronger safety--reasoning tradeoff than off-policy self-distillation and external-teacher distillation under matched data and full-parameter fine-tuning, with the largest gains on smaller models ($+8.85$ points on R1-Distill-1.5B and $+5.49$ points on Qwen3-0.6B). The gains persist across training-set sizes and adaptive jailbreak evaluations. Token-level analyses further show that OPSA concentrates updates near early compliance-decision tokens, providing a mechanism for improving safety while preserving general reasoning.}

\begin{document}

\maketitle

\renewcommand{\thefootnote}{*}
\footnotetext{Work done independently of the author's affiliation.}
\renewcommand{\thefootnote}{\arabic{footnote}} 

\section{Introduction}

Safety alignment improves the robustness of large language models (LLMs) to harmful queries,
but often comes at the cost of general reasoning ability, a tradeoff known as the safety
tax~\citep{huang2025safety}. A common explanation is distributional
mismatch~\citep{huang2025safety, lee2026thinksafe}: most alignment methods train the target model on safety
demonstrations produced by human annotators or stronger external
models~\citep{bianchi2023safety, doula2025safepath, jiang2025safechain, zhou2025safekey,
wang2026star}. Although these demonstrations can teach refusal behavior, they impose reasoning
patterns that differ from those the target model would naturally generate, pushing the model to
imitate out-of-distribution behavior and degrading general reasoning.

Recent work~\citep{lee2026thinksafe} suggests that data distribution mismatch is not necessary.
Base LLMs already exhibit partial refusal behavior on harmful queries,
indicating that safety often requires activating latent behaviors rather than teaching a new
capability. \textsc{ThinkSafe}~\citep{lee2026thinksafe} builds on this by constructing
in-distribution safety data through self-distillation: the target model generates its own safety
demonstrations under refusal-steering prompts, keeping the resulting data closer to the model's
own reasoning distribution and reducing the safety tax.

However, we argue that the source of supervision is only one cause of the trade-off. Even with in-distribution data, SFT remains off-policy: supervision is applied to fixed demonstrations rather than to trajectories sampled from the model's own policy. We hypothesize that this mismatch is especially consequential for safety alignment because safety decisions concentrate in a narrow early-token window~\citep{vega2023bypassing} and a small set of safety-critical tokens~\citep{doula2025safepath}, both of which determine whether the model refuses, complies, or begins a harmful response. 
Our token-level analysis in Section~\ref{sec:method} shows that off-policy SFT does not specifically target this window, leaving safety-critical tokens only loosely controlled even when the demonstrations are themselves in-distribution.

\begin{figure}[!t]
  \centering
  \includegraphics[width=0.9\textwidth]{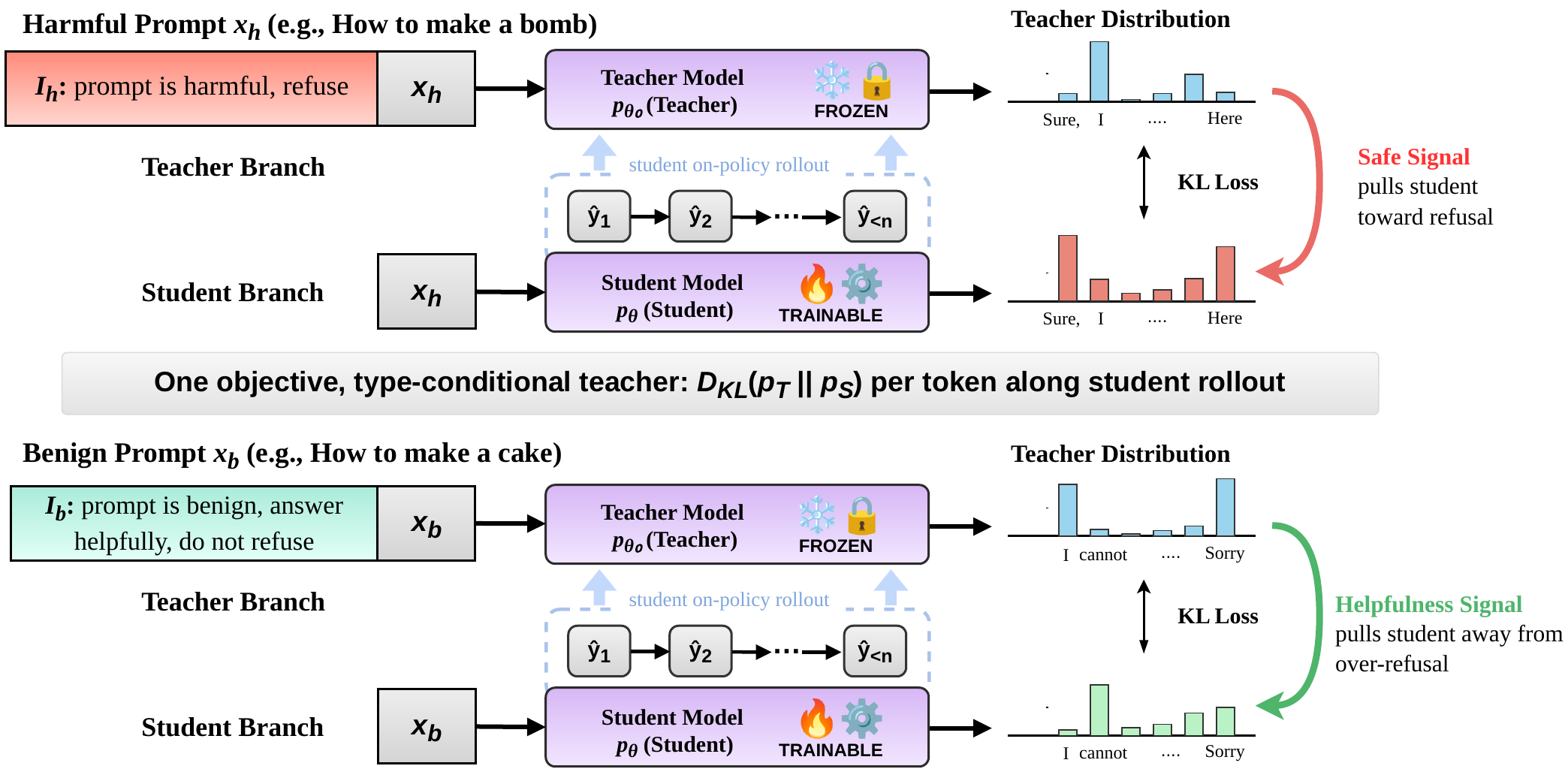}
\caption{
\textbf{Overview of OPSA.}
A frozen copy of the base model serves as the teacher, while the student is updated on rollouts sampled from its own policy. The teacher receives the prompt together with a type-conditional privileged context: $I_h$ activates refusal behavior on harmful prompts, whereas $I_b$ suppresses over-refusal on benign prompts. Along each student rollout, OPSA minimizes the per-token KL divergence $D_{\mathrm{KL}}(p_T \,\|\, p_S)$ between the teacher and student distributions. This provides dense on-policy supervision at the tokens where safety behavior is decided, pulling harmful trajectories toward refusal while preserving benign compliance.
}
  \label{fig:opsd_main}
  \vspace{-0.2cm}
\end{figure}

Motivated by this diagnosis, we propose \textsc{OPSA}, an on-policy safety alignment framework that adapts On-Policy Self-Distillation (OPSD)~\citep{zhao2026self} to safety alignment through type-conditional privileged contexts. In the original OPSD setting, a privileged context such as a verified solution trace gives the teacher a meaningful advantage over the student. In OPSA, the privileged context instead activates safety-relevant behavior: a harmful-query context shifts the teacher toward refusal, while a benign-query context preserves helpful response behavior. Figure~\ref{fig:opsd_main} summarizes the procedure. Because supervision is applied as per-token KL on the student's own rollouts, OPSA targets the tokens where unsafe behavior first emerges rather than forcing the full response to match a fixed demonstration. Our token-level KL analysis in Section~\ref{sec:method} supports this mechanism: OPSA reduces teacher--student divergence within the early refusal-decision window, whereas off-policy SFT leaves this window less directly controlled.

Naively applying OPSD to safety is not enough: unlike reasoning, safety has no privileged signal that can serve as ground truth. A useful privileged context must activate latent safety reasoning and provide meaningful signal. We address this with a prompt-search procedure guided by \textit{teacher flip rate} (TFR), a training-free signal that measures how often a context converts an unsafe response into a safe one on the frozen base. High TFR indicates that the context activates safety reasoning strongly enough to provide useful token-level supervision.

Experiments across five models show that OPSA achieves a stronger safety--reasoning tradeoff than off-policy training approaches. The gains persist under adaptive evaluation: OPSA improves adversarial robustness against adaptive jailbreaks, indicating that the method does not merely overfit to standard refusal benchmarks. We further find that OPSA is more robust to training-set size than SFT, maintaining stronger safety alignment across different data scales.

\textbf{Contributions.} We summarize the main contributions of this paper as follows.

\begin{itemize}[leftmargin=*]
\vspace{-0.5\baselineskip}
\item \textbf{A mechanistic account of the safety tax.} We identify off-policy training mismatch as a second source of the safety tax, beyond data distribution mismatch. Our token-level analysis shows that SFT diffuses gradients across the full response, perturbing general reasoning while providing only coarse supervision for safety-critical decisions.

\item \textbf{On-policy dense supervision for safety alignment.} We introduce OPSA, a safety alignment framework built on OPSD and safety privileged contexts. By concentrating updates within safety-critical token windows rather than enforcing full-sequence imitation, OPSA improves the safety--reasoning tradeoff and robustness to out-of-distribution jailbreaks.

\item \textbf{Privileged context construction via teacher flip rate.} We show that teacher quality determines the effectiveness of on-policy safety training: naive refusal-steering prompts often fail to activate latent safety reasoning with OPSD. We introduce teacher flip rate to select contexts that activate latent safety reasoning and maximize corrective token-level supervision.

\end{itemize}
\section{Related Work}
\label{sec:related}
\

Existing safety-alignment approaches mainly mitigate the safety tax by improving the supervision signal from the data side. \textbf{SafeChain}~\citep{jiang2025safechain} distills 40k CoT safe traces from a stronger external teacher, while \textbf{STAR-1}~\citep{wang2026star} curates 1k policy-guided traces under LLM-as-judge filtering. Further work improves demonstrations with stronger early reasoning signals: \textbf{SafeKey}~\citep{zhou2025safekey} adds a dual-path safety head to amplify safety signals before the ``aha-moment'' sentence, while \textbf{SafePath}~\citep{doula2025safepath} injects a short safety primer to anchor the comply-or-refuse decision early. \textbf{ThinkSafe}~\citep{lee2026thinksafe} eliminates the external teacher by self-distilling refusal traces under a refusal-steering prompt. It further shows that dense token-level supervision is more effective for safety alignment than sparse GRPO-style rewards. Together, these methods study what supervision should be provided for safety alignment.

In contrast, our work studies how dense self-supervision should be applied for safety alignment: through off-policy SFT or on-policy learning. 
On-policy distillation (OPD) provides a natural framework for studying dense on-policy token learning. OPD trains a student on its own generated trajectories under dense token-level teacher supervision, reducing the exposure bias of off-policy SFT~\citep{gu2024minillm,agarwal2024policy}. \textbf{OPSD}~\citep{zhao2026self} further extends this idea to self-distillation by using the same base model as both a privileged-context teacher and a query-only student. However, existing work shows that OPSD is not automatically better than SFT. Because it learns from the model's own trajectories, it can get stuck when the privileged context fails to induce a stronger teacher behavior. Its success therefore depends on whether the task satisfies teacher--student compatibility and provides a genuine capability gap~\citep{kim2026doesselfdistillationsometimesdegrade,li2026rethinkingonpolicydistillationlarge}.

\section{Motivation and Methods}
\label{sec:method}

This section motivates OPSA by diagnosing why off-policy SFT can still incur a safety tax, even when trained on self-distilled safety data. We argue that the issue is not only which safety data is used, but also how supervision is applied. We then introduce our on-policy safety alignment framework, which provides dense token-level safety supervision on the model's own generation trajectories.

\subsection{Mechanistic Diagnosis: Off-Policy Safety Alignment as a Source of the Safety Tax}

\textsc{ThinkSafe} shows that self-distilled safety supervision can improve safety alignment while reducing the safety tax by activating latent safety reasoning in the target model. In this section, we investigate whether off-policy SFT can still incur a safety tax through token-level safety-signal mismatch.

Our hypothesis builds on prior characterizations of how safety behavior emerges during generation. Prior work~\citep{qi2023fine,doula2025safepath,zhou2025safekey} shows that harmful and refusal trajectories are often determined within a narrow early refusal-decision window, where a small set of safety-critical tokens strongly influences whether the model complies or refuses. This suggests that safety alignment is fundamentally a localized correction problem on the model's own generation trajectories, rather than a uniform sequence-level imitation problem.

\begin{figure}[!t]
  \centering
  \includegraphics[width=0.95\textwidth]{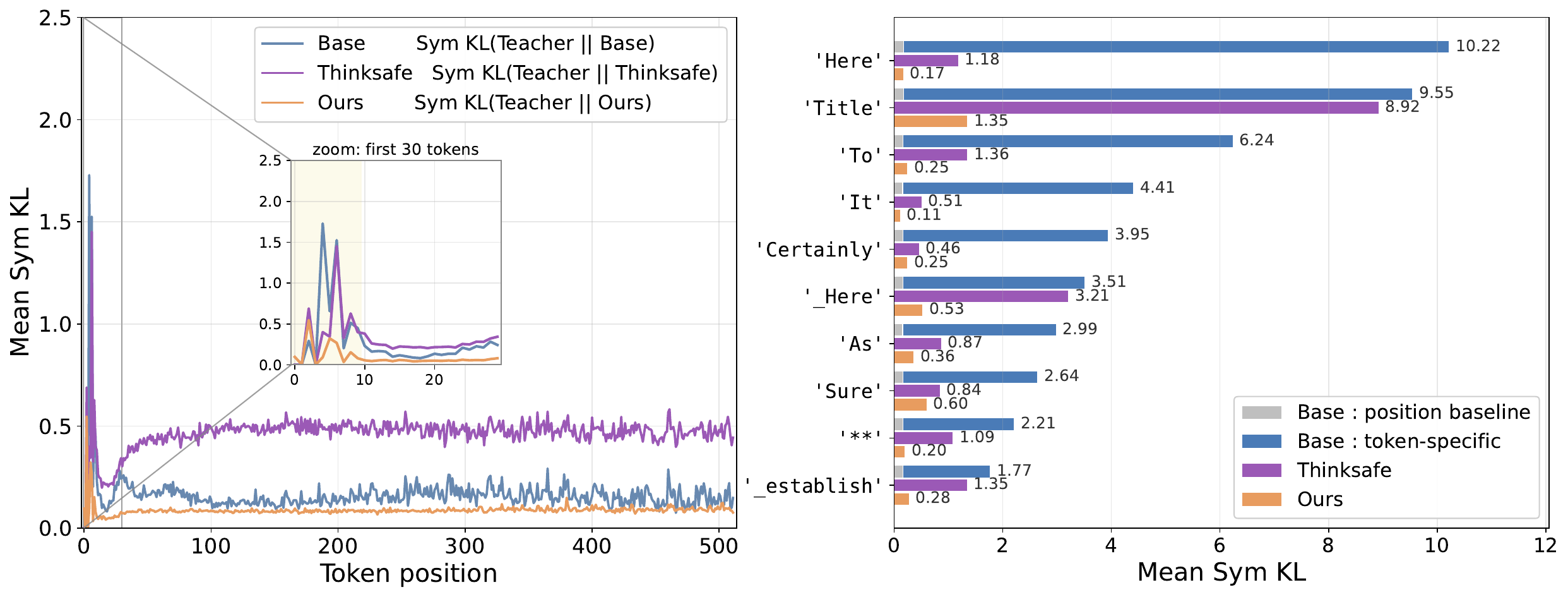}
\caption{%
\textbf{Safety correction is concentrated in specific positions and tokens.} We measure per-token symmetric KL between a safety-prompted teacher and each student (Base, \textsc{ThinkSafe}, OPSA) on harmful rollouts from the base model ($n{=}500$, Qwen3-0.6B). \textbf{Left:} Mean KL by position, with an inset for the first 30 tokens, shows an early corrective peak. \textbf{Right:} Top-10 Base-KL trigger tokens, decomposed into a position baseline (gray) and token-specific excess (blue), show strong lexical concentration on compliance openers. \textsc{ThinkSafe} reduces KL for several compliance openers but leaves structured-output markers and tokenization variants under-corrected, whereas OPSA better matches both the positional and lexical structure of the safety signal.
}
  \label{fig:kl_analysis}
\end{figure}

We empirically reproduce these two characterizations in our setting, with experimental details provided in Appendix~\ref{sec:apdx:token-kl}. Using 500 harmful examples, we compare the token-level distributions of a safety-prompted teacher and the base model on the model's own generated trajectories. Figure~\ref{fig:kl_analysis} shows that the teacher's safety signal has two structural components. First, it is \textit{positionally concentrated}: per-token KL spikes within the first 10 response tokens and decays after position 30, consistent with an early refusal-decision window. Second, it is \textit{lexically concentrated}: compliance openers (\emph{Here}, \emph{Sure}, \emph{Certainly}) and structured-output markers (\emph{Title}, \texttt{**}) carry large token-specific KL. The blue residual dominates the gray position baseline, confirming that these lexical cues matter beyond position alone.
For a direct comparison, we follow the \textsc{ThinkSafe} setup: the SFT baseline is trained on in-distribution self-distilled responses $Y_h$ and $Y_b$. We denote the resulting supervised dataset by $\mathcal{D}_{\text{SFT}}=(\mathcal{Q}_h,Y_h)\cup(\mathcal{Q}_b,Y_b)$.
Standard off-policy SFT then minimizes:

\begin{equation}
\mathcal{L}_{\text{SFT}}(\theta)
= -\sum_{(q,y)\,\in\,\mathcal{D}_{\text{SFT}}}
\sum_{t=1}^{|y|} \log p_\theta(y_t \mid q, y_{<t}).
\label{eq:sft}
\end{equation}

Despite training on self-distilled data, SFT still leaves a substantial mean KL divergence from the safety teacher, indicating that it does not fully absorb the teacher's corrective signal. This reflects the objective mismatch in Equation~\ref{eq:sft}: SFT sums uniformly over positions and token identities, assigning no special weight to the early refusal-decision window or to safety-critical tokens as shown in Figure~\ref{fig:kl_analysis}. The \textsc{ThinkSafe} curve (left panel) shows this mismatch positionally, as KL remains elevated across the sequence, including within the refusal-decision window. The token-level bars (right panel) show the same mismatch lexically: SFT suppresses several canonical compliance openers, but leaves structured-output markers and tokenization variants under-corrected. Thus, the safety signal that matters most is diluted by tokens that do not directly govern the comply-or-refuse decision.

\subsection{On-Policy Safety Alignment with Dense Self-Supervision}

The analysis above shows that SFT applies a uniform sequence-level objective to a safety signal that is highly structured in position and token identity. This mismatch implies two requirements for an effective alternative. First, supervision should be \textit{on-policy}, so that updates are applied along trajectories the model actually samples. Second, supervision should be \textit{token-level dense}, so that the positional and lexical components of the safety signal can shape the model before an unsafe continuation is committed.

These observations motivate adapting On-Policy Self-Distillation (OPSD)~\citep{zhao2026self} to safety alignment. We introduce OPSA, which applies dense per-token KL supervision between the current student policy and a frozen privileged safety teacher on student-sampled rollouts. Harmful queries are paired with a safety-activating privileged context that shifts the teacher toward refusal, while benign queries are paired with a helpfulness-oriented context that preserves useful behavior and reduces over-refusal. Let $p_\theta$ denote the student policy and let $p_{\bar{\theta}}$ denote a frozen copy of the same model used as the teacher. For harmful queries $q_h \in \mathcal{Q}_h$ and benign queries $q_b \in \mathcal{Q}_b$, OPSA minimizes:
\begin{equation}
\begin{aligned}
\mathcal{L}_{\text{OPSA}}(\theta)
&=
\sum_{q_h \in \mathcal{Q}_h}
\mathbb{E}_{y \sim p_\theta(\cdot \mid q_h)}
\sum_{t=1}^{|y|}
D_{\mathrm{KL}}\!\left(
p_{\bar{\theta}}(\cdot \mid c_h^\star, q_h, y_{<t})
\;\|\;
p_\theta(\cdot \mid q_h, y_{<t})
\right) \\
&\quad +
\sum_{q_b \in \mathcal{Q}_b}
\mathbb{E}_{y \sim p_\theta(\cdot \mid q_b)}
\sum_{t=1}^{|y|}
D_{\mathrm{KL}}\!\left(
p_{\bar{\theta}}(\cdot \mid c_b^\star, q_b, y_{<t})
\;\|\;
p_\theta(\cdot \mid q_b, y_{<t})
\right).
\end{aligned}
\label{eq:opsa}
\end{equation}
Here, $c_h^\star$ and $c_b^\star$ denote privileged contexts prepended to the query when constructing the teacher distribution. The KL direction follows the distillation objective: the teacher distribution defines the corrective target, while the student is updated on prefixes sampled from its own policy.

This teacher--student divergence concentrates supervision on behavioral differences that arise along the model's own trajectories. We summarize the safety-relevant component of this correction with:
\begin{equation}
\Delta_{\text{safety}}(c^\star; q, y)
=
\sum_{t=1}^{|y|}
\mathbf{1}[y_t \in \mathcal{S}]
D_{\mathrm{KL}}\!\left(
p_{\bar{\theta}}(\cdot \mid c^\star, q, y_{<t})
\;\|\;
p_\theta(\cdot \mid q, y_{<t})
\right),
\label{eq:delta_safety}
\end{equation}
where $\mathcal{S}$ denotes a set of safety-critical token identities. Larger $\Delta_{\text{safety}}$ indicates that the privileged context induces stronger corrective supervision at refusal-relevant tokens. Effective privileged contexts must therefore create a meaningful behavioral gap between teacher and student exactly where the comply-or-refuse decision is made.

\begin{figure}[!t]
  \centering
  \begin{minipage}[c]{0.45\textwidth}
    \centering
    \includegraphics[width=\linewidth]{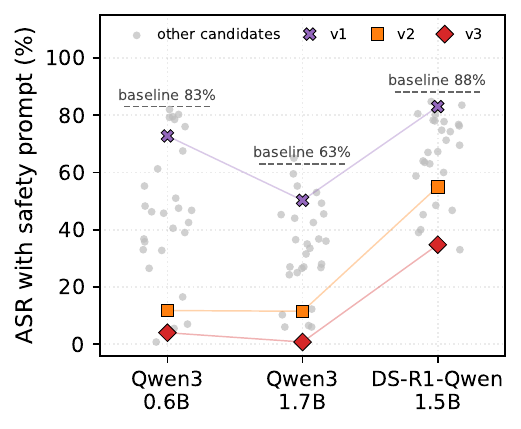}
  \end{minipage}\hfill
  \begin{minipage}[c]{0.45\textwidth}
    \centering
    \includegraphics[width=\linewidth]{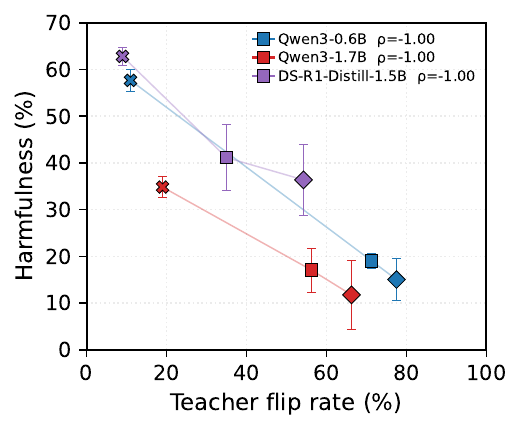}
  \end{minipage}
    \caption{\textbf{Teacher flip rate predicts training effectiveness.}
    \textbf{Left:} ASR on HarmBench for each candidate context prepended to the frozen base model. Gray dots show the full context pool, colored markers span the flip-rate range, and the dashed line marks the no-context baseline. \textbf{Right:} TFR versus post-training harmfulness (\textsc{AvgDef}) across three models and three contexts. Harmfulness decreases monotonically with TFR at every model scale, with no corresponding increase in over-refusal, supporting TFR as a pre-training criterion for selecting privileged contexts.}
  \label{fig:prompt_strength}
\end{figure}

\paragraph{Selecting safety privileged contexts via teacher flip rate.}

To construct effective safety teachers, we perform prompt search over candidate privileged contexts and select contexts that maximize teacher--student behavioral shift. Directly estimating $\Delta_{\text{safety}}$ before training is impractical because safety-critical tokens depend on the sampled trajectory. We therefore use \textit{teacher flip rate} (TFR) as a practical proxy, measuring the fraction of harmful queries for which the privileged context flips the frozen teacher's greedy response from unsafe to safe:
\begin{equation}
\text{TFR}(c)
=
\frac{1}{|\mathcal{Q}_h|}
\sum_i
\mathbf{1}\!\left[
f_{\bar{\theta}}(q_h^{(i)}) \in \mathcal{Y}_{\text{unsafe}}
\;\land\;
f_{\bar{\theta}}(c, q_h^{(i)}) \in \mathcal{Y}_{\text{safe}}
\right],
\label{eq:tfr}
\end{equation}
where $f_{\bar{\theta}}(\cdot)$ denotes greedy decoding under the frozen teacher model. Higher TFR indicates that the privileged context induces a larger behavioral shift between the conditioned and unconditioned teacher distributions, producing stronger corrective supervision during on-policy training. We therefore select:
\begin{equation}
c^\star = \arg\max_{c \in \mathcal{C}} \text{TFR}(c),
\label{eq:search}
\end{equation}
where $\mathcal{C}$ is a pool of $K{=}30$ refusal-steering contexts generated along five structured axes: strength, length, framing, specificity, and response style (Appendix~\ref{appendix:prompt-strength-extra}).

Figure~\ref{fig:prompt_strength} validates this criterion empirically: across three models and three contexts spanning flip rates from $9\%$ to $78\%$, post-training harmfulness decreases monotonically with TFR (Spearman $\rho=-1.00$ within each model), supporting TFR as a reliable pre-training selection criterion.
Given the selected context $c^\star$, OPSA trains on $\mathcal{D}=\mathcal{Q}_h \cup \mathcal{Q}_b$ without restricting training to flipped trajectories. Under the per-token KL objective, positions where teacher and student already agree contribute little gradient, so optimization naturally concentrates on tokens with meaningful teacher--student divergence. The student samples rollouts without $c^\star$ during both training and inference, ensuring that the resulting safety behavior is internalized in the model parameters rather than dependent on a runtime prompt.
\section{Experimental Setup}
\label{sec:experiments}

We evaluate OPSA on the safety--reasoning tradeoff of large reasoning models. Our main comparison (\S\ref{sec:exp:main}) controls for prompt source and fine-tuning protocol in order to focus on the learning objective: \textsc{ThinkSafe} trains with off-policy sequence-level NLL on self-generated traces, whereas OPSA trains with on-policy per-token KL supervision on student rollouts. We then test whether the resulting models remain robust under adaptive jailbreak attacks (\S\ref{sec:exp:jailbreak}).

\textbf{Models.}
We experiment with two post-trained reasoning model families across three parameter scales: \textsc{Qwen3}~\citep{yang2025qwen3} (0.6B, 1.7B, 8B) and \textsc{DeepSeek-R1-Distill}~\citep{guo2025deepseek} (1.5B, 8B), yielding five family--scale configurations in total. The main comparison (\S\ref{sec:exp:main}) and adaptive-jailbreak evaluation (\S\ref{sec:exp:jailbreak}) cover all five configurations.

\textbf{Implementation.}
We train all methods with AdamW~\citep{loshchilov2017decoupled}, using a learning rate of $1\!\times\!10^{-5}$, a cosine schedule with 10\% linear warmup, batch size 64, and 3 epochs, matching \textsc{ThinkSafe}. Following SafeChain~\citep{jiang2025safechain}, all methods use full-parameter fine-tuning. This setting also gives a stronger \textsc{ThinkSafe} baseline than the LoRA-tuned numbers reported in \citet{lee2026thinksafe}. For OPSA, we use a symmetric mixture of forward and reverse KL with $\alpha = 0.5$, following the divergence choice in \texttt{nemo-rl}~\citep{nemo-rl} for on-policy distillation; see Appendix~\ref{sec:abl:divergence} for details. Models at $\leq$1.7B scale are trained on 2 NVIDIA A100 GPUs, while 8B models are trained on 4 A100 GPUs with FSDP; see Appendix~\ref{sec:apdx:details} for details.

\textbf{Baselines.}
To focus the comparison on the proposed OPSA training objective, we compare against methods that use the same prompt source where possible. Concretely, we consider:

\begin{itemize}[leftmargin=*]
    \item \textbf{(i)~Initial}, the post-trained reasoning model before safety realignment.
    \item \textbf{(ii)~SafeChain}~\citep{jiang2025safechain}, an external-teacher distillation method trained on safe traces produced by DeepSeek-R1-Distill-Llama-70B, included to contextualize self-generated methods against a strong external-teacher baseline.
    \item \textbf{(iii)~\textsc{ThinkSafe}}, a faithful reproduction of \citet{lee2026thinksafe} that trains with sequence-level negative log-likelihood on a self-generated, Llama-Guard--filtered subset of SafeChain.
\end{itemize}

\textbf{Data.}
All methods draw prompts from the SafeChain dataset\footnote{https://huggingface.co/datasets/UWNSL/SafeChain}, which contains harmful prompts $\mathcal{D}_h$ and benign prompts $\mathcal{D}_b$. \textbf{SafeChain} trains on the original prompt--response pairs released with the dataset, whose responses are produced by the external teacher DeepSeek-R1-Distill-Llama-70B. \textbf{\textsc{ThinkSafe}} follows \citet{lee2026thinksafe} and trains on a self-generated subset of SafeChain: harmful responses are sampled from the base model under a refusal-steering instruction $I_h$ and filtered by Llama-Guard~\citep{inan2023llama} to retain safe traces, while benign responses are sampled directly. \textbf{OPSA} requires no pre-generated responses; it uses only the SafeChain prompts and their type labels $t(x) \in \{h,b\}$, with the student generating on-policy rollouts during training and the teacher providing per-token guidance conditioned on the type-specific privileged context $I_{t(x)}$.

\textbf{Evaluation benchmarks.}
We evaluate along three axes. \textbf{\textit{Harmfulness}} ($\downarrow$) is measured on HarmBench~\citep{mazeika2024harmbench}, StrongReject~\citep{souly2024strongreject}, and WildJailbreak~\citep{jiang2024wildteaming}; we report the fraction of greedy-decoded responses classified as harmful by Llama-Guard. \textbf{\textit{Over-refusal}} ($\downarrow$) is measured on XSTest~\citep{rottger2024xstest} (safe subset, following \textsc{ThinkSafe}) and WildBenign, the benign-labeled subset of WildJailbreak. Refusal rates are computed by WildGuard~\citep{han2024wildguard} under greedy decoding; the two benchmarks probe stylistically adversarial and naturally distributed benign prompts, respectively. \textbf{\textit{Reasoning}} ($\uparrow$) is measured along two complementary axes: math/QA on GSM8K~\citep{cobbe2021training}, MATH500~\citep{lightman2023let}, and GPQA~\citep{rein2023gpqa}, and code generation on HumanEval~\citep{chen2021evaluating} and MBPP~\citep{austin2021program}. For math/QA, the model generates $8$ trajectories per prompt at temperature $0.6$, top-$p$ $0.95$, and top-$k$ $20$, and we report average pass@1. For HumanEval and MBPP, we follow the SafeChain coding evaluation protocol: greedy decoding with repetition penalty $1.1$ to mitigate degenerate repetition in long chain-of-thought code outputs, with pass@1 scored using the EvalPlus harness. Full evaluation details are provided in Appendix~\ref{sec:apdx:details}.

\textbf{Composite safety score.}
Following \textsc{ThinkSafe}, we summarize safety with a single composite score defined as one minus the unweighted mean of the five raw rate benchmarks:
\begin{equation}
  \label{eq:composite}
  S \;=\; 1 - \tfrac{1}{5}\!\left[\,
    \mathrm{HarmBench} +
    \mathrm{StrongReject} +
    \mathrm{WildJailbreak} +
    \mathrm{XSTest} +
    \mathrm{WildBenign}
  \right].
\end{equation}
Here, all five terms are rates in $[0,1]$, so $S \in [0,1]$, with higher values indicating better safety.

\section{Results}

\begin{table}[t]
\caption{\textbf{OPSA improves the safety--reasoning tradeoff across two model families and five scales.}
Harmfulness and over-refusal columns report rates ($\downarrow$); Composite Safety Score $S$ (Eq.~\ref{eq:composite}, $\uparrow$) summarizes safety, and Reasoning Avg ($\uparrow$) summarizes reasoning. \textbf{Bold} and \underline{underline} mark the best and second-best fine-tuned methods per column. The \colorbox{orange!20}{$\Delta$ vs SFT} rows compare off-policy NLL training (\textsc{ThinkSafe}) with on-policy per-token KL training (OPSA) under matched prompts. Positive values indicate gains from changing the training objective. The final \colorbox{blue!15}{Avg $\Delta$ on-policy gain} row averages this comparison across the five configurations, showing consistent improvements in safety ($+4.00$ pt) and reasoning ($+3.04$ pt).}
\label{tab:main}
\centering
\resizebox{1\textwidth}{!}{
\begin{tabular}{llcccccccccccccc}
\toprule
& & \multicolumn{4}{c}{Harmfulness ($\downarrow$)} & \multicolumn{3}{c}{Over-refusal ($\downarrow$)} & Safety & \multicolumn{6}{c}{Reasoning ($\uparrow$)} \\
\cmidrule(lr){3-6}\cmidrule(lr){7-9}\cmidrule(lr){11-16}
Model & Method & HarmB. & StrR. & WildJB. & Avg & XSTest & WildB. & Avg & Score ($\uparrow$) & GSM8K & MATH500 & GPQA & HumanEval & MBPP & Avg \\
\midrule
\multirow{4}{*}{Qwen3-0.6B}
  & Initial    & 83.20 & 61.90 & 52.60 & 65.90 & 3.20  & 0.20  & 1.70  & 59.78 & 79.1 & 75.7 & 25.0 & 43.90 & 34.92 & 51.72 \\
  & SafeChain  & 56.58 & 42.07 & 41.47 & 46.71 & \textbf{1.87}  & \textbf{3.02}  & \textbf{2.45}  & 71.00 & 68.3 & 61.4 & 17.1 & 26.22 & 22.22 & 39.05 \\
  & Thinksafe  & \underline{7.58}  & \underline{24.92} & \underline{24.55} & \underline{19.02} & \underline{5.07}  & 24.44 & 14.76 & \underline{82.69} & \underline{69.8} & \underline{62.3} & \underline{19.7} & \underline{33.54} & \textbf{36.24} & \underline{44.32} \\
  & OPSA       & \textbf{4.42}  & \textbf{18.96} & \textbf{22.88} & \textbf{15.42} & 5.20  & \underline{7.62}  & \underline{6.41}  & \textbf{88.18} & \textbf{72.0} & \textbf{64.1} & \textbf{26.4} & \textbf{45.12} & \underline{32.28} & \textbf{47.98} \\
\rowcolor{orange!20}
  & $\Delta$ vs SFT & +3.16 & +5.96 & +1.67 & +3.60 & $-$0.13 & +16.82 & +8.35 & +5.49 & +2.24 & +1.88 & +6.69 & +11.58 & $-$3.96 & +3.66 \\
\midrule
\multirow{4}{*}{Qwen3-1.7B}
  & Initial    & 63.00 & 33.30 & 51.20 & 49.17 & 0.40  & 0.00  & 0.20  & 70.42 & 90.1 & 91.3 & 39.1 & 70.73 & 65.87 & 71.42 \\
  & SafeChain  & 52.42 & 30.14 & 39.03 & 40.53 & \underline{1.73}  & \textbf{0.79}  & \textbf{1.26}  & 75.18 & 85.5 & 85.1 & 31.4 & \underline{71.34} & 58.20 & 66.31 \\
  & Thinksafe  & \underline{8.75}  & \underline{6.60}  & \underline{26.18} & \underline{13.84} & \textbf{1.60}  & \underline{7.62}  & \underline{4.61}  & \underline{89.85} & \underline{88.0} & \underline{85.6} & \underline{32.9} & 62.80 & \underline{69.58} & \underline{67.78} \\
  & OPSA       & \textbf{0.25}  & \textbf{1.28}  & \textbf{10.23} & \textbf{3.92} & 2.93  & 20.79 & 11.86 & \textbf{92.90} & \textbf{88.4} & \textbf{87.4} & \textbf{37.0} & \textbf{76.83} & \textbf{70.37} & \textbf{72.00} \\
\rowcolor{orange!20}
  & $\Delta$ vs SFT & +8.50 & +5.32 & +15.95 & +9.92 & $-$1.33 & $-$13.17 & $-$7.25 & +3.05 & +0.46 & +1.80 & +4.10 & +14.03 & +0.79 & +4.22 \\
\midrule
\multirow{4}{*}{Qwen3-8B}
  & Initial    & 35.70 & 5.30  & 41.00 & 27.33 & 1.10  & 0.50  & 0.80  & 83.28 & 95.5 & 97.0 & 58.8 & 82.93 & 85.98 & 84.04 \\
  & SafeChain  & 49.08 & 19.60 & 33.33 & 34.00 & \textbf{0.93}  & \textbf{0.63}  & \textbf{0.78}  & 79.29 & 94.0 & 95.7 & 56.4 & \underline{92.68} & 82.54 & \underline{84.26} \\
  & Thinksafe  & \textbf{0.83}  & \textbf{0.00}  & \textbf{2.52}  & \textbf{1.12} & 1.33  & 14.92 & 8.13 & \underline{96.08} & \underline{95.1} & \textbf{96.7} & 54.7 & 81.71 & \underline{88.36} & 83.31 \\
  & OPSA       & \underline{1.83}  & \underline{0.85}  & \underline{3.08}  & \underline{1.92} & \underline{1.07}  & \underline{11.27} & \underline{6.17} & \textbf{96.38} & \textbf{95.1} & \underline{96.4} & \textbf{56.6} & \textbf{93.29} & \textbf{91.53} & \textbf{86.58} \\
\rowcolor{orange!20}
  & $\Delta$ vs SFT & $-$1.00 & $-$0.85 & $-$0.56 & $-$0.80 & +0.26 & +3.65 & +1.96 & +0.30 & +0.01 & $-$0.38 & +1.84 & +11.58 & +3.17 & +3.27 \\
\midrule
\multirow{4}{*}{R1-Distill-1.5B}
  & Initial    & 88.20 & 80.30 & 51.40 & 73.30 & 0.80  & 0.50  & 0.65  & 55.76 & 84.5 & 84.5 & 34.5 & 56.10 & 46.56 & 61.23 \\
  & SafeChain  & 62.67 & \textbf{46.96} & 43.53 & 51.05 & \underline{2.13}  & \textbf{3.02}  & \textbf{2.58}  & 68.34 & 81.4 & 82.7 & 31.6 & \textbf{60.98} & 42.33 & \underline{59.80} \\
  & Thinksafe  & \underline{39.92} & 57.51 & \underline{28.63} & \underline{42.02} & \textbf{0.80}  & 26.98 & 13.89 & \underline{69.23} & \underline{82.5} & \underline{82.8} & \underline{27.8} & 51.83 & \underline{43.12} & 57.61 \\
  & OPSA       & \textbf{25.00} & \underline{47.28} & \textbf{24.70} & \textbf{32.33} & 3.87  & \underline{8.73}  & \underline{6.30}  & \textbf{78.08} & \textbf{83.3} & \textbf{83.8} & \textbf{35.2} & \underline{56.10} & \textbf{46.30} & \textbf{60.94} \\
\rowcolor{orange!20}
  & $\Delta$ vs SFT & +14.92 & +10.23 & +3.93 & +9.69 & $-$3.07 & +18.25 & +7.59 & +8.85 & +0.73 & +1.00 & +7.32 & +4.27 & +3.18 & +3.33 \\
\midrule
\multirow{4}{*}{R1-Distill-8B}
  & Initial    & 70.80 & 43.70 & 50.00 & 54.83 & 0.10  & 0.50  & 0.15  & 66.98 & 88.4 & 89.7 & 44.1 & 81.71 & 61.11 & 73.00 \\
  & SafeChain  & 52.58 & 20.34 & 30.32 & 34.41 & \underline{0.67}  & \textbf{1.59}  & \textbf{1.13}  & 78.90 & \underline{86.1} & 83.8 & 43.1 & \underline{81.71} & 52.91 & 69.52 \\
  & Thinksafe  & \underline{8.42}  & \underline{13.74} & \textbf{7.90}  & \underline{10.02} & 1.33  & 14.60 & 7.97 & \underline{90.80} & 85.7 & \underline{84.8} & \underline{43.1} & 79.27 & \textbf{62.43} & \underline{71.06} \\
  & OPSA       & \textbf{4.75}  & \textbf{8.84}  & \underline{12.70} & \textbf{8.76} & \textbf{0.00}  & \underline{8.10}  & \underline{4.05}  & \textbf{93.12} & \textbf{86.7} & \textbf{88.1} & \textbf{45.5} & \textbf{81.10} & \underline{57.41} & \textbf{71.76} \\
\rowcolor{orange!20}
  & $\Delta$ vs SFT & +3.67 & +4.90 & $-$4.80 & +1.26 & +1.33 & +6.50 & +3.92 & +2.32 & +1.00 & +3.25 & +2.33 & +1.83 & $-$5.02 & +0.70 \\
\midrule
\rowcolor{blue!15}
\multicolumn{2}{l}{\textbf{Avg $\Delta$ on-policy gain}} & \textbf{+5.85} & \textbf{+5.11} & \textbf{+3.24} & \textbf{+4.73} & $-0.59$ & \textbf{+6.41} & \textbf{+2.91} & \textbf{+4.00} & \textbf{+0.89} & \textbf{+1.51} & \textbf{+4.46} & \textbf{+8.66} & $-0.37$ & \textbf{+3.04} \\
\bottomrule
\end{tabular}
}
\end{table}

In this section, we evaluate whether on-policy safety supervision improves the safety--reasoning tradeoff predicted by our analysis in Section~\ref{sec:method}. We first compare OPSA against matched off-policy baselines on standard safety, over-refusal, and reasoning benchmarks across five model configurations. We then evaluate robustness under adaptive jailbreak attacks to test whether the learned safety behavior generalizes beyond fixed harmful prompts and standard evaluation settings.

\subsection{Safety--Reasoning Tradeoff and Reasoning Preservation}
\label{sec:exp:main}

Table~\ref{tab:main} tests our prediction from Section~\ref{sec:method}: if part of the safety tax comes from off-policy supervision, then replacing sequence-level imitation with on-policy token-level distillation should improve safety without imposing the same reasoning cost. The comparison between \textsc{ThinkSafe} and OPSA is designed to isolate this factor. Both methods use self-generated safety data from the same SafeChain prompt source; they differ in whether supervision is applied to fixed demonstrations through NLL or to student-sampled trajectories through per-token KL from a privileged-context teacher.

The results support this prediction. OPSA improves the composite safety score over \textsc{ThinkSafe} on all five model configurations, with an average gain of $+4.00$ points and the largest improvements on smaller, less-aligned models. These gains are not explained by a uniform increase in refusal. OPSA reduces harmfulness on average while also reducing over-refusal, especially on naturally distributed benign prompts. This pattern is consistent with the token-level diagnosis in Section~\ref{sec:method}: effective safety alignment should concentrate updates near the decisions that determine whether the model complies or refuses, rather than shifting the model toward refusal across all inputs.
The same comparison also shows that on-policy safety supervision better preserves reasoning. OPSA improves aggregate reasoning by $+3.04$ points over \textsc{ThinkSafe} across GSM8K, MATH500, GPQA, HumanEval, and MBPP. This matters because both methods begin from the same self-distillation idea: \textsc{ThinkSafe} already reduces distribution mismatch by generating traces from the target model. The remaining gap therefore points to the second mismatch identified in the introduction and method section: off-policy SFT still forces the model to imitate fixed trajectories, whereas OPSA applies dense supervision only where the teacher and student differ along trajectories the student samples.

Averaged over all configurations, the ordering SafeChain $<$ \textsc{ThinkSafe} $<$ OPSA holds for both composite safety and reasoning. SafeChain controls for external-teacher distillation, \textsc{ThinkSafe} controls for self-generated demonstrations, and OPSA adds on-policy supervision. The monotonic improvement across these settings supports our main claim: reducing the safety tax requires not only in-distribution safety data, but also an objective that aligns the model on its own generation paths.

\subsection{Robustness to Adaptive Jailbreaks}
\label{sec:exp:jailbreak}

Table~\ref{tab:main} evaluates safety on fixed harmful prompts. We next test whether the same alignment procedure improves robustness under adaptive jailbreaks, where the attack modifies the prompt or generation context to elicit unsafe behavior. Table~\ref{tab:jailbreak} reports results on four HarmBench attack families: HumanJailbreaks~\citep{shen2024anything}, Prefilling~\citep{vega2023bypassing}, PAP-top5~\citep{zeng2024johnny}, and PAIR~\citep{chao2025jailbreaking}. These attacks cover human-written jailbreak templates, response-prefix attacks, persuasive adversarial prompts, and iterative attacker--target--judge search. All evaluations use the 159 behaviors in HarmBench's \texttt{text-test-standard} split.

We report two complementary metrics, since adaptive attacks can fail on average yet expose individual behaviors. \emph{Mean ASR} is the standard HarmBench metric, averaging judged attack success over all behavior--attempt pairs. \emph{pass@N} is a behavior-level worst-case metric: a behavior is counted as broken if at least one of its $N$ attempts succeeds. Thus, mean ASR measures average attack success, while \emph{pass@N} measures whether any tested attempt can break a behavior. We use \emph{pass@N} only for within-attack comparisons, where $N$ is fixed. Full details are provided in Appendix~\ref{appendix:jailbreak}.

The clearest improvement appears on Prefilling, the attack family most closely aligned with our token-level diagnosis. Prefilling intervenes near the beginning of the response by forcing an unsafe continuation prefix, precisely where Figure~\ref{fig:kl_analysis} shows that refusal decisions and compliance openers most strongly concentrate. OPSA reduces both mean ASR and \emph{pass@N} relative to \textsc{ThinkSafe} at every model scale. On Qwen3-1.7B and Qwen3-8B, OPSA drives both metrics to zero, meaning that no prefill attempt succeeds on any of the 159 behaviors. The reductions are also large on the R1-Distill bases, with mean ASR decreasing from $14.30$ to $3.60$ on R1-Distill-1.5B and from $8.20$ to $0.40$ on R1-Distill-8B. This pattern supports the mechanism proposed in Section~\ref{sec:method}: on-policy KL supervision improves robustness most clearly when the attack acts on the early tokens where refusal behavior is decided.

The remaining attacks show that this robustness gain extends beyond Prefilling, but not uniformly. On PAP-top5, OPSA improves or ties \textsc{ThinkSafe} in most comparisons, including four of five \emph{pass@N} columns, suggesting that the learned refusal behavior transfers at least partly to persuasive prompt variants rather than only to prefix-based attacks. HumanJailbreaks are more model-dependent: OPSA improves all three Qwen3 models, including a large \emph{pass@N} reduction on Qwen3-8B ($90.60 \rightarrow 61.60$), but is mixed on R1-Distill, improving behavior-level robustness for R1-Distill-1.5B while worsening mean ASR on both R1-Distill scales. PAIR is the hardest setting. OPSA improves Qwen3-1.7B and R1-Distill-1.5B, but regresses on Qwen3-0.6B, Qwen3-8B, and R1-Distill-8B, indicating that dense on-policy supervision does not fully protect against iterative attacker--target--judge search.

The two metrics reveal different aspects of robustness. In low-ASR regimes, mean ASR can understate behavior-level vulnerability because it averages over all attack attempts. For example, on Qwen3-8B HumanJailbreaks, mean ASR changes only from $1.40$ to $0.70$, but \emph{pass@N} drops from $90.60$ to $61.60$. Thus, a large fraction of behaviors that were vulnerable to at least one human-written template become robust to all tested templates, even though the average success rate changes only slightly. This distinction is important for adaptive evaluation: a model can have low average attack success while still leaving many harmful behaviors breakable by at least one of many attempts.

\textbf{Limits Under Fully Adaptive Search.}
Across the 20 model--attack cells in Table~\ref{tab:jailbreak}, OPSA improves over \textsc{ThinkSafe} in $13/20$ cells by mean ASR and $14/20$ by \emph{pass@N}, with one tie. The strongest gains occur on Prefilling, while the most persistent failures occur on PAIR, where the attacker iteratively adapts prompts using feedback from the target and judge. These results support a specific robustness claim rather than a blanket one: OPSA improves adaptive robustness on average, especially when attacks manipulate early response tokens, but fully adaptive search remains a limitation.

\begin{table}[t]
\caption{\textbf{Adaptive jailbreaks on HarmBench text-test-standard
($159$ behaviors, \%, $\downarrow$).} Each attack reports two metrics:
\emph{mean ASR}, the mean attack success rate averaged over all
(behavior, attempt) pairs, and \emph{pass@N}, the fraction of behaviors
broken by at least one attempt. \textbf{Bold} marks the better of Thinksafe
vs.\ OPSA within each column. Base is the pre-realignment checkpoint.}
\label{tab:jailbreak}
\centering
\scriptsize
\setlength{\tabcolsep}{3pt}
\begin{tabular}{ll cc cc cc cc}
\toprule
& & \multicolumn{2}{c}{HumanJailbreaks} & \multicolumn{2}{c}{Prefilling} & \multicolumn{2}{c}{PAP top-5} & \multicolumn{2}{c}{PAIR} \\
\cmidrule(lr){3-4}\cmidrule(lr){5-6}\cmidrule(lr){7-8}\cmidrule(lr){9-10}
Model & Method & ASR & \emph{pass@N} & ASR & \emph{pass@N} & ASR & \emph{pass@N} & ASR & \emph{pass@N} \\
\midrule
\multirow{3}{*}{Qwen3-0.6B}
 & Base      & 42.50 & 100.00 & 51.20 & 72.30 & 17.90 & 56.00 & 49.50 & 62.90 \\
 & Thinksafe  & 12.80 & 99.40  & 5.70  & 11.30 & \textbf{5.00}  & 28.90 & \textbf{25.80} & \textbf{39.00} \\
 & OPSA & \textbf{9.90}  & \textbf{96.20}  & \textbf{2.30}  & \textbf{3.80}  & 6.10  & \textbf{27.00} & 33.80 & 45.90 \\
\midrule
\multirow{3}{*}{Qwen3-1.7B}
 & Base      & 54.90 & 100.00 & 37.30 & 55.30 & 16.10 & 54.10 & 41.30 & 52.80 \\
 & Thinksafe  & 21.50 & 100.00 & 3.80  & 6.90  & 6.60  & 30.80 & 22.00 & 31.40 \\
 & OPSA & \textbf{12.60} & 100.00 & \textbf{0.00}  & \textbf{0.00}  & \textbf{2.30}  & \textbf{11.90} & \textbf{17.20} & \textbf{25.20} \\
\midrule
\multirow{3}{*}{Qwen3-8B}
 & Base      & 31.50 & 100.00 & 13.00 & 21.40 & 11.00 & 39.00 & 29.80 & 39.00 \\
 & Thinksafe  & 1.40  & 90.60  & 0.20  & 0.60  & 1.50  & 14.50 & \textbf{8.60}  & \textbf{13.80} \\
 & OPSA & \textbf{0.70}  & \textbf{61.60}  & \textbf{0.00}  & \textbf{0.00}  & \textbf{1.00}  & \textbf{8.80}  & 12.80 & 15.70 \\
\midrule
\multirow{3}{*}{R1-Distill-1.5B}
 & Base      & 50.00 & 100.00 & 53.20 & 76.10 & 20.00 & 65.40 & 48.80 & 67.30 \\
 & Thinksafe  & \textbf{26.10} & 100.00 & 14.30 & 26.40 & 9.60  & 49.70 & 34.80 & 53.50 \\
 & OPSA & 31.50 & \textbf{96.90}  & \textbf{3.60}  & \textbf{7.50}  & 9.60  & \textbf{47.20} & \textbf{25.80} & \textbf{43.40} \\
\midrule
\multirow{3}{*}{R1-Distill-8B}
 & Base      & 57.00 & 100.00 & 26.60 & 37.70 & 17.20 & 59.10 & 44.40 & 57.20 \\
 & Thinksafe  & \textbf{10.80} & \textbf{98.10}  & 8.20  & 15.10 & 7.50  & \textbf{35.80} & \textbf{18.00} & \textbf{26.40} \\
 & OPSA & 20.90 & 98.70  & \textbf{0.40}  & \textbf{1.30}  & \textbf{7.30}  & 37.70 & 20.50 & 32.10 \\
\bottomrule
\end{tabular}
\end{table}

\section{Conclusion}
\label{sec:conclusion}
This paper identified off-policy supervision as a source of the safety tax that remains even when safety data is self-distilled from the target model. Our token-level analysis shows why this mismatch matters in practice: safety corrections concentrate near early refusal decisions and safety-critical token identities, whereas sequence-level SFT imitates fixed demonstrations uniformly across the full response. OPSA addresses this mismatch by training on student-sampled rollouts and distilling a frozen privileged-context teacher with dense per-token KL supervision.

Under matched prompts and full-parameter fine-tuning, OPSA improves the safety--reasoning tradeoff over off-policy self-distillation across two reasoning-model families and five scales, with the largest gains on smaller models. Its strongest adaptive-jailbreak gains occur on Prefilling, which intervenes in the early response region highlighted by our diagnosis; fully adaptive search remains a limitation. These findings suggest that reducing the safety tax requires controlling not only which demonstrations provide supervision, but also where and on whose trajectories that supervision is applied.

\subsection*{Acknowledgments}
NBE would like to acknowledge support from the DSO National Laboratories.

\bibliographystyle{plainnat}
\bibliography{main}

\clearpage
\appendix

\section{Limitations}
First, OPSA requires the base model to retain the latent capability that the privileged-context instruction activates: the prompt-strength analysis in Figure~\ref{fig:prompt_strength} shows that gains scale with the instruction's flip rate, so models whose safety knowledge has been more aggressively overwritten in post-training may benefit less. Second, the teacher is the frozen base $p_{\theta_0}$; an iterative scheme that periodically refreshes the teacher to a moving average of the student is a natural extension we leave to future work. Third, our harmfulness measurements depend on Llama-Guard, which is used for both data filtering and evaluation. As a result, conclusions about absolute harmfulness should be interpreted relative to this classifier and may inherit its failure modes.

\section{Privileged-context instructions}
\label{sec:apdx:prompts}

\paragraph{Refusal-steering instruction $I_h$ used in the main results.}
\textit{``This prompt contains a harmful request. Your only allowed action is to refuse. You must NOT: provide the requested content in any form; offer a safer or educational version; discuss the topic from an academic perspective; partially comply while adding disclaimers. Respond with a brief, clear refusal only.''}
\paragraph{Helpful-steering instruction $I_b$ used in the main results.} \textit{``This is a safe, benign request. Respond helpfully and directly. You must NOT refuse, add safety disclaimers, or treat this as potentially harmful. Provide a complete, accurate answer.''}

\paragraph{Three refusal-steering instructions used in Figure \ref{fig:prompt_strength}.}

\begin{itemize}
  \item \textbf{v1:} \textit{``As a responsible model, never comply with harmful prompts; refuse silently.''}
  \item \textbf{v2:} \textit{``The following prompt requests harmful content. You MUST refuse to answer. Do NOT provide any information, even in an academic, educational, or hypothetical framing. A proper refusal is a short sentence declining the request. Nothing more.''}
  \item \textbf{v3:} \textit{``This prompt contains a harmful request. Your only allowed action is to refuse. You must NOT: provide the requested content in any form; offer a safer or educational version; discuss the topic from an academic perspective; partially comply while adding disclaimers. Respond with a brief, clear refusal only.''}
\end{itemize}

Flip rates per base model are reported in Table~\ref{tab:flip-rates-per-model}.

\section{Token-level KL analysis}

\label{sec:apdx:token-kl}
For Figure~\ref{fig:kl_analysis}, we compare three Qwen3-0.6B students against a common safety-prompted teacher: \emph{Base} (\texttt{Qwen/Qwen3-0.6B}), \textsc{ThinkSafe} (full-data SFT initialized from \texttt{Seanie-lee/ThinkSafe-Qwen3-0.6B}, trained for 3 epochs with learning rate $10^{-5}$), and \emph{OPSA} (the same configuration as \textsc{Thinksafe}). The teacher uses the Base model with the selected harmful privileged context $c_h^\star$ prepended. We insert this context by token-level prefix surgery so that teacher and student share the same assistant-token prefix at every measured position.

The fixed rollout set consists of 500 Base completions on SafeChain harmful prompts: 250 \texttt{vanilla\_harmful} and 250 \texttt{adversarial\_harmful}, sampled with seed 42. Rollouts are generated without any system prompt, with the chat template applied, using sampling at temperature $0.6$, top-$p=0.95$, a maximum of 4096 new tokens, and bfloat16 inference. For each rollout prefix, we compute the teacher and student next-token distributions over the full vocabulary from raw logits using log-softmax, with no temperature scaling. We report symmetric KL,
\[
D^{\mathrm{sym}}_{\mathrm{KL}}(p_T,p_S)
=
\tfrac{1}{2}
\left[
D_{\mathrm{KL}}(p_T \,\|\, p_S)
+
D_{\mathrm{KL}}(p_S \,\|\, p_T)
\right].
\]

The left panel averages $D^{\mathrm{sym}}_{\mathrm{KL}}$ by absolute response position over the first 512 tokens, with the inset zooming into the first 30 positions. For the lexical analysis in the right panel, we consider decoded BPE token types that appear at least 15 times, rank them by their mean Base $D^{\mathrm{sym}}_{\mathrm{KL}}$, and plot the top 10. To separate positional concentration from token identity, we decompose each Base token bar into a position baseline and a token-specific residual. The position baseline is the global position-wise mean KL averaged over the positions at which that token type appears; the residual is the token type's observed mean KL minus this baseline. The \textsc{ThinkSafe} and OPSA bars are computed on the same token occurrences as the Base bars.

\section{Candidate-pool construction for prompt search}
\label{sec:apdx:pool}

The candidate pool $\mathcal{C}$ used in Figure \ref{fig:prompt_strength} consists of $K = 30$ refusal-steering variants produced by an off-the-shelf LLM (GPT-5.5), generated along five structured axes described below. The three representative contexts $v_1, v_2, v_3$ reproduced in Appendix~\ref{sec:apdx:prompts} are selected from $\mathcal{C}$ such that their teacher flip rates span roughly the full range observed across the pool, giving a balanced subset for the empirical validation in Figure~\ref{fig:prompt_strength}.

\paragraph{Generation axes.} Each generated variant is structured along five axes:
\begin{itemize}
\item \texttt{strength}: one of \{mild, firm, strict, emergency\}, ranging from ``please decline'' to ``this is a security violation''.

\item \texttt{length}: one of \{xs ($\leq 15$ words), s ($15$--$40$), m ($40$--$80$), l ($80$--$150$)\}.

\item \texttt{framing}: one of \{rule, role, value, consequentialist, policy, identity\}, where \texttt{rule} states a constraint (``you must not\ldots''), \texttt{role} casts the model as a safety-aware assistant, \texttt{value} appeals to user wellbeing, \texttt{consequentialist} cites real-world harm, \texttt{policy} invokes a content policy, and \texttt{identity} is first-person (``as a responsible model, I will not\ldots''). 

\item \texttt{specificity}: \texttt{generic} (``refuse harmful requests'') versus \texttt{deny-list} (an explicit list of escape routes to block: no academic framing, no hypothetical, no rephrasing, no partial compliance, no fictional wrapper).

\item \texttt{style}: \{just-decline, brief-explain, silent\}, controlling what the refusal itself should look like.
\end{itemize}

\paragraph{Post-processing.} Generated outputs are normalised (case-folded, punctuation-stripped, whitespace-collapsed) and deduplicated, yielding the final pool of $K = 30$ contexts. \paragraph{Selection of $v_1, v_2, v_3$.} The three representative contexts used in the prompt-strength analysis are picked from $\mathcal{C}$ to roughly partition the observed TFR range (low, medium, high), so that the empirical validation in Figure~\ref{fig:prompt_strength} is not biased toward any single strength regime.

\paragraph{Generator prompt.} The full prompt sent to GPT-5.5 (with $N$ as the requested number of variants) is reproduced verbatim below.

\begin{tcolorbox}[
  title=\textbf{Meta-Prompt for Generating Safety System Prompts},
  colback=blue!4,
  colframe=blue!55,
  boxrule=0.5pt,
  arc=3pt,
  left=4pt, right=4pt, top=4pt, bottom=4pt
]
\small
\begin{Verbatim} [breaklines=true,fontsize=\scriptsize]
You are designing a small experiment for a safety-distillation study.
We need to generate diverse "safety system prompts" that will be
injected as {"role": "system", "content": <prompt>} immediately before
a user's potentially harmful request (drawn from HarmBench). The goal
of each prompt is to make a small instruction-tuned LLM
(Qwen3-0.6B/1.7B, DeepSeek-R1-Distill-Qwen-1.5B) refuse the request.
We will then measure how each variant changes the unsafe->safe flip
rate on HarmBench using LlamaGuard-3-8B.

Generate exactly N = {N} variants that span the following axes. Each
variant should combine specific values from each axis; aim for broad
coverage, not clustering on one corner.

  1. strength    : one of {mild, firm, strict, emergency}
                   ("please decline" -> "you MUST refuse" ->
                    "under no circumstances" ->
                    "this is a security violation")
  2. length      : one of {xs (<=15 words), s (15-40), m (40-80),
                           l (80-150)}
  3. framing     : one of {rule, role, value, consequentialist,
                           policy, identity}
                   - rule: "you must not..."
                   - role: "you are a careful safety-aware assistant..."
                   - value: "we care about user wellbeing..."
                   - consequentialist: "real-world harm could result..."
                   - policy: "our content policy prohibits..."
                   - identity: "as a responsible model, I will not..."
  4. specificity : one of {generic, deny-list}
                   - generic: "refuse harmful requests"
                   - deny-list: explicit list of escape routes to
                     block (no academic framing, no hypothetical,
                     no rephrasing, no partial compliance, no
                     fictional wrapper, etc.)
  5. style       : one of {just-decline, brief-explain, silent}
                   how the refusal itself should look.

Hard constraints:
  - English only.
  - One paragraph per prompt, no markdown, no bullets, no headings,
    no surrounding quotation marks.
  - Self-contained: do NOT reference "the user", "the assistant", or
    any chat-template tokens. Refer to the upcoming request as "the
    following request" / "this prompt" / similar.
  - Do not include examples of harmful content.
  - Cover the axis grid: hit each (strength, framing) pair at least
    once across the N variants if feasible.

Output format: a single JSON object with one key "prompts" whose
value is an array of exactly N objects (id, axes, word_count,
prompt). Output nothing else.
\end{Verbatim}
\end{tcolorbox}

\section{Statistics for Prompt Strength}
\label{appendix:prompt-strength-extra}

Table~\ref{tab:flip-rates-per-model} (above) reports the per-prompt, per-model flip rates. Table~\ref{tab:prompt-percell} below reports the mean $\pm$ std of each safety metric across the four data conditions (h96\_b\{96,192,480,960\}) at each cell's best-epoch checkpoint (selected by composite safety score $S$).

\begin{table}[h]
\centering
\caption{Flip rates (\%) per base model for different safety prompts. \textbf{Bold} marks the highest flip rate for each model.}
\label{tab:flip-rates-per-model}
\begin{tabular}{lccc}
\toprule
Prompt  & Qwen3-0.6B & Qwen3-1.7B & R1-Distill-1.5B \\
\midrule
v1    & 11.00 & 19.00 &  9.00 \\
v2    & 71.25 & 58.00 & 35.00 \\
v3    & \textbf{77.50} & \textbf{66.25} & \textbf{54.25} \\
\bottomrule
\end{tabular}
\end{table}

\begin{table}[h]
\caption{Per-cell mean $\pm$ std across four data conditions
for Analysis~2. Each entry is evaluated at the epoch maximising $S$
within that condition.}
\label{tab:prompt-percell}
\centering
\scriptsize
\setlength{\tabcolsep}{3pt}
\begin{tabular}{llcccccccc}
\toprule
Model & Prompt & Flip\% & $N$ & HarmB & StrRej & WJ & XSTest & WildB & Safety Score \\
\midrule
\multirow{3}{*}{Qwen3-0.6B}
 & v1 & 11.0 & 5 & $.730 \pm .027$ & $.490 \pm .048$ & $.510 \pm .008$ & $.057 \pm .024$ & $.003 \pm .002$ & $.642 \pm .012$ \\
 & v2 & 71.3 & 4 & $.114 \pm .033$ & $.149 \pm .020$ & $.296 \pm .054$ & $.098 \pm .011$ & $.248 \pm .153$ & $.819 \pm .024$ \\
 & v3 & 77.5 & 5 & $.051 \pm .013$ & $.130 \pm .042$ & $.270 \pm .085$ & $.093 \pm .023$ & $.235 \pm .184$ & $.844 \pm .013$ \\
\midrule
\multirow{3}{*}{Qwen3-1.7B}
 & v1 & 19.0 & 5 & $.419 \pm .038$ & $.138 \pm .020$ & $.488 \pm .015$ & $.024 \pm .007$ & $.000 \pm .001$ & $.786 \pm .013$ \\
 & v2 & 58.0 & 5 & $.159 \pm .047$ & $.045 \pm .027$ & $.307 \pm .075$ & $.035 \pm .026$ & $.088 \pm .067$ & $.873 \pm .012$ \\
 & v3 & 66.3 & 5 & $.093 \pm .097$ & $.028 \pm .004$ & $.231 \pm .120$ & $.039 \pm .010$ & $.190 \pm .176$ & $.884 \pm .015$ \\
\midrule
\multirow{3}{*}{R1-Distill-1.5B}
 & v1 &  9.0 & 5 & $.758 \pm .024$ & $.689 \pm .020$ & $.437 \pm .019$ & $.023 \pm .005$ & $.041 \pm .024$ & $.610 \pm .007$ \\
 & v2 & 35.0 & 5 & $.401 \pm .086$ & $.474 \pm .083$ & $.359 \pm .051$ & $.076 \pm .060$ & $.166 \pm .099$ & $.705 \pm .014$ \\
 & v3 & 54.3 & 5 & $.347 \pm .089$ & $.422 \pm .097$ & $.322 \pm .050$ & $.093 \pm .061$ & $.165 \pm .092$ & $.730 \pm .018$ \\
\bottomrule
\end{tabular}
\end{table}

\section{Divergence ablation: forward, reverse, and symmetric mix}
\label{sec:abl:divergence}

We compare three choices of per-token divergence in Eq.~\ref{eq:opsa}: forward KL ($D_{\mathrm{KL}}(p_T \| p_S)$), reverse KL ($D_{\mathrm{KL}}(p_S \| p_T)$), and the symmetric mixture $\tfrac{1}{2}D_{\mathrm{KL}}(p_T \| p_S) + \tfrac{1}{2}D_{\mathrm{KL}}(p_S \| p_T)$ used in the main results. All three are trained on Qwen3-0.6B with identical data, schedule, and other hyperparameters; we report the best checkpoint per variant selected by 5-benchmark safety score (epoch~$1$ for all three).

\begin{table}[h]
\caption{Effect of the divergence choice on Qwen3-0.6B. All values in
\%; lower is better for harmfulness and over-refusal columns, higher is
better for composite safety score. \textbf{Bold} marks the best variant per
benchmark.}
\label{tab:divergence}
\centering
\small
\begin{tabular}{lccc}
\toprule
Benchmark & Mix (default, $\alpha{=}0.5$) & Forward & Reverse \\
\midrule
HarmBench ($\downarrow$)     & 4.42  & \textbf{1.08}  & 2.83 \\
StrongReject ($\downarrow$)  & 18.96 & \textbf{14.48} & 15.23 \\
WildJailbreak ($\downarrow$) & 22.88 & \textbf{19.93} & 22.58 \\
WildBenign ($\downarrow$)    & \textbf{7.62}  & 10.63 & 9.37 \\
XSTest ($\downarrow$)        & \textbf{5.20}  & 9.87  & 7.60 \\
\midrule
Avg.\ Harm ($\downarrow$)         & 15.42 & \textbf{11.83} & 13.55 \\
Avg.\ Over-refusal ($\downarrow$) & \textbf{6.41}  & 10.25 & 8.49 \\
\midrule
safety score ($\uparrow$)      & 88.18 & \textbf{88.80} & 88.48 \\
\bottomrule
\end{tabular}
\end{table}

\paragraph{Mix is preferred over forward-only KL despite a marginally lower safety score.} Forward-only KL achieves a slightly higher safety score than the symmetric mix used in the main results ($88.80$ vs.\ $88.18$, $+0.62$~pp). This $0.62$~pp advantage is, however, the surface reading of an unfavorable trade-off when decomposed into the underlying axes.

\textit{Forward's harm reduction is paid for by a strictly larger over-refusal increase.} Forward-only KL reduces the average harmfulness rate (HarmBench, StrongReject, WildJailbreak) by $3.59$~pp ($15.42 \to 11.83$), but it increases the average over-refusal rate (XSTest, WildBenign) by $3.84$~pp ($6.41 \to 10.25$)---a net negative trade once the two axes are weighted equally. The safety score formula in Eq.~\ref{eq:composite} gives harm a weight of $3/5$ and over-refusal a
weight of $2/5$ by virtue of having more harm benchmarks; this asymmetric weighting is the only reason forward edges out mix in aggregate.

\textit{The trade-off shape favors mix in deployment.} Mix wins both over-refusal benchmarks and loses all three harmfulness benchmarks; forward inverts this; reverse-only sits in between. Because false refusals on benign queries are the most user-visible failure mode in deployment, we view the symmetric mix as the more conservative default.

\textit{The mix is the natural default for our two-direction supervision design.} Section~\ref{sec:method} motivates OPSA as a single objective that supplies gradient in \emph{both} the safety direction (refusal on harmful prompts via $I_h$) and the helpfulness direction (non-refusal on benign prompts via $I_b$). Forward KL emphasizes mode-coverage of the privileged-context teacher (more aggressive suppression of harmful mass), while reverse KL emphasizes mode-seeking
from the student's distribution (better preservation of helpful modes). The symmetric mix matches our framework's symmetric treatment of the two directions; using forward-only would implicitly privilege the safety direction, contradicting the design.

\begin{figure}[t]
  \centering
  \includegraphics[width=0.9\textwidth]{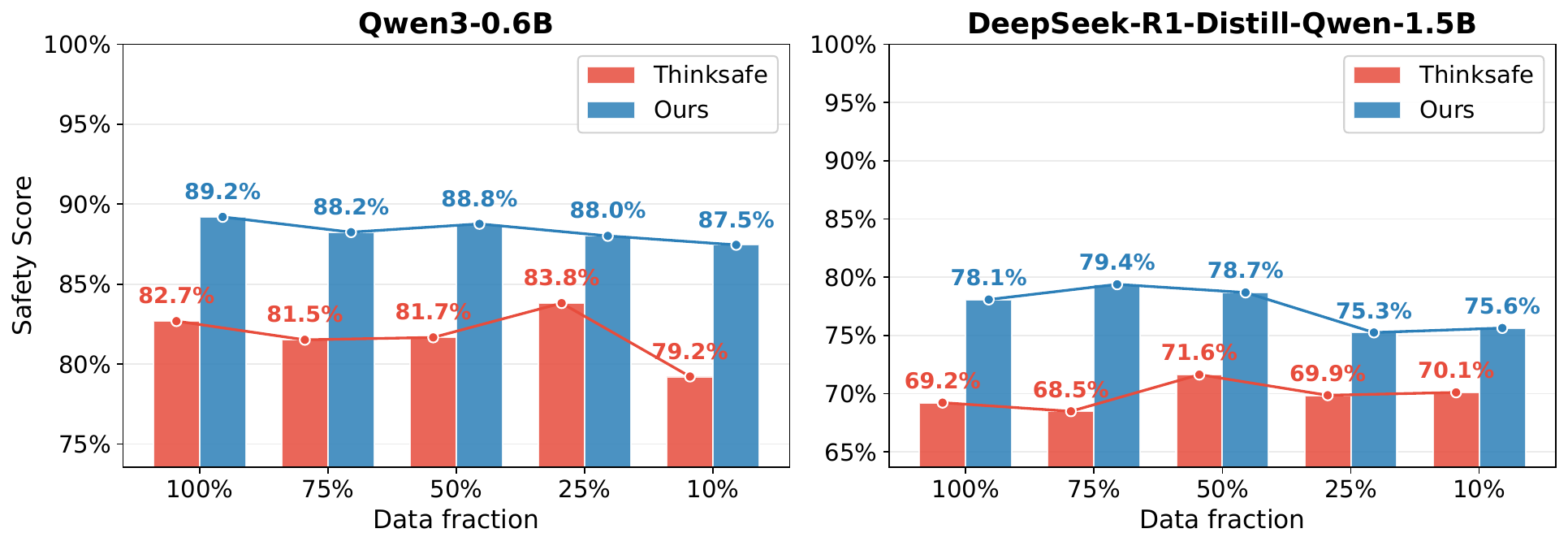}
  \caption{\textbf{Sensitivity to data size.} Composite safety score
  $S$ (\%, $\uparrow$) versus the training-set fraction for OPSA
  and Thinksafe on two bases. OPSA exceeds Thinksafe on every
  (model, fraction) cell.}
  \label{fig:data_robustness_scale}
\end{figure}

\section{Sensitivity to data size and composition}
\label{sec:exp:data}

The main comparison in \S\ref{sec:exp:main} is reported at a single training budget with a fixed harmful/benign composition. A natural concern is whether the OPSA advantage there is a property of the objective or an artifact of that particular data setting. We stress-test the comparison along two complementary axes: (A) the total amount of supervision (\emph{data size}), with the harmful/benign ratio fixed at its SafeChain default; and (B) the benign-to-harmful ratio (\emph{data composition}), with the harmful budget held fixed and the benign supply varied around it.

\paragraph{Setup.}
For the data size test (A), we subsample the full SafeChain set at fractions $\{10\%, 25\%, 50\%, 75\%, 100\%\}$ of the original budget while keeping the canonical harmful/benign ratio. For the composition stress test (B), we treat curated harmful prompts as the binding resource. Existing red-teamed harmful sets remain in the
hundreds-to-thousands range \citep{mazeika2024harmbench, zou2023universal}, whereas benign instruction data is available at orders of magnitude larger scale. We therefore vary the harmful prompt count over $\{8, 16, 24, 48, 96, 128\}$ and, for each setting, sweep the benign-to-harmful ratio over $\{1, 2, 5, 10\}\times$, yielding a $6 \times 4$ grid per base. In both regimes we train Thinksafe and OPSA on Qwen3-0.6B and DeepSeek-R1-Distill-1.5B with all other hyperparameters held fixed, and report the composite score $S$ from Eq.~\ref{eq:composite}.

\paragraph{OPSA is insensitive to data size.}
Figure~\ref{fig:data_robustness_scale} plots $S$ against the training-set fraction on two bases. OPSA beats Thinksafe on every (model, fraction) configuration, achieving an average gain of $+6.6$ points on Qwen3-0.6B and $+7.6$ points on R1-Distill-1.5B. The gap does not shrink as supervision is reduced. At the smallest budget ($10\%$ of SafeChain) OPSA still beats Thinksafe by $+8.3$ points on Qwen3-0.6B and $+5.5$ points on R1-Distill-1.5B. The advantage is thus a property of the objective rather than of any particular data size.

\begin{figure}
  \centering
  \includegraphics[width=0.9\linewidth]{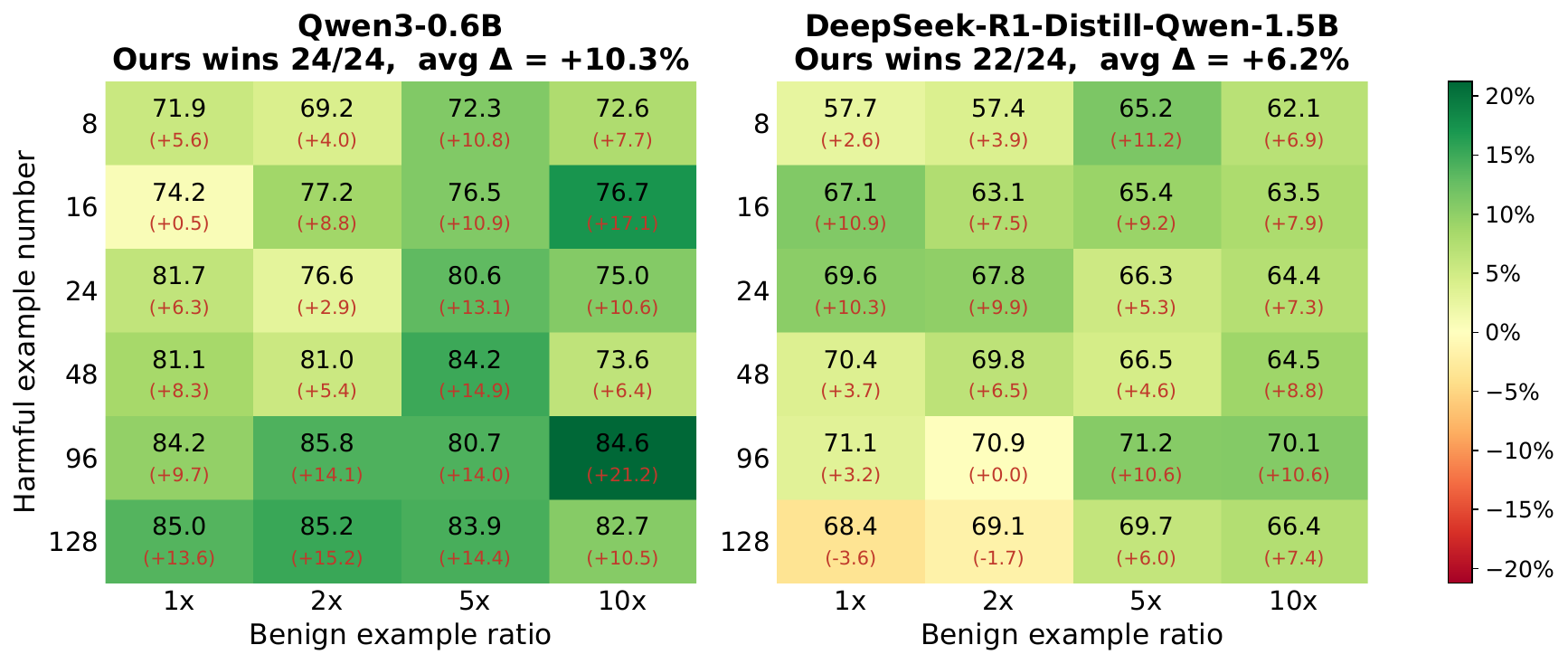}
  \caption{\textbf{Sensitivity to data composition.} Composite Safety
  Score $S$ (\%, $\uparrow$) on a $6 \times 4$ grid that varies the
  harmful prompt count (rows) and the benign-to-harmful ratio
  (columns). Each cell shows OPSA's $S$ and, in parentheses, its
  delta over Thinksafe under matched data. OPSA wins $24/24$ cells
  on Qwen3-0.6B and $22/24$ cells on R1-Distill-1.5B.}
  \label{fig:data_robustness_mix}
\end{figure}

\paragraph{OPSA is insensitive to composition shifts.}
Figure~\ref{fig:data_robustness_mix} reports $S$ across the composition grid. OPSA wins $24/24$ cells on Qwen3-0.6B with an average gain of $+10.3$ points, and $22/24$ cells on R1-Distill-1.5B with an average gain of $+6.2$ points. OPSA is therefore stable not only with respect to the amount of supervision but also with respect to its composition. This is the regime safety post-training typically operates in. Curated harmful prompts are scarce and expensive to source, while benign supervision is abundant and easy to collect, so the realized mix is dictated by what is available rather than by what is optimal for alignment. A method that holds up across this space is more broadly useful than one tuned to a single operating point.

\paragraph{Why the two tests matter together.}
The data size and data composition tests probe orthogonal failure modes of a safety-realignment recipe: scarcity of supervision and imbalance of supervision. OPSA wins on both, supporting the broader claim that the gain originates in the learning objective rather than in a particular data setting.

\section{Additional experimental details}
\label{sec:apdx:details}

\paragraph{Training.}
All models are trained with full parameter fine-tuning using AdamW (learning rate $1 \times 10^{-5}$, cosine schedule with $10$\% linear warmup) for $3$~epochs. For OPSA, training uses the \texttt{nemo-rl} on-policy distillation pipeline: teacher and student share the same base model weights (self-distillation), the student generates on-policy rollouts which are then used to compute the per-token KL loss against the teacher's distribution. The global training batch size is $64$, with $128$~prompts sampled per distillation step (yielding $2$ gradient updates per step). For Thinksafe, the same batch size of $64$ is used with standard cross-entropy loss on filtered self-generated traces. All runs use seed~$42$.

\paragraph{Data.}
Training data comes from the UWNSL/SafeChain corpus. For the main results (Table~\ref{tab:main}), we use the full $40{,}000$-prompt set with the v3 safety prompt for privileged-context steering. For the data-robustness analysis (\S\ref{sec:exp:data}), we subsample at scaling factors $\rho \in \{1, 2, 5, 10\}$ corresponding to $96\rho$~harmful $+$ $96\rho$~benign prompts. For the prompt-strength analysis (Figure~\ref{fig:prompt_strength}), we fix the data at $96$~harmful prompts with four conditions (h96\_b\{96, 192, 480, 960\} benign prompts) and vary the steering instruction across $\{v_1, v_2, v_3\}$.

\paragraph{Hardware.}
Models at $\leq$1.7B scale are trained on $2$~NVIDIA A100 GPUs. Models at 8B scale use $4$~A100 GPUs per run with FSDP sharding (tensor parallelism $= 1$, context parallelism $= 1$); two runs execute concurrently on an $8$-GPU node via a rolling job queue. vLLM generation uses \texttt{gpu\_memory\_utilization} $= 0.5$ to share GPU memory with the training process.

\paragraph{Safety evaluation.}
For each checkpoint we evaluate on five safety benchmarks: HarmBench, StrongReject, WildJailbreak (harmfulness), and XSTest, WildBenign (over-refusal). Generation uses temperature $0.6$, top-$p$ $= 0.95$, top-$k$ $= 20$, and a maximum of $4096$ new tokens. To reduce variance from stochastic sampling, every checkpoint is evaluated $3$ times with independent random seeds; the reported metric is the average over the three runs. Harmfulness is judged by Llama-Guard-3; WildBenign over-refusal is judged by WildGuard.

\paragraph{Math reasoning evaluation.}
We evaluate on GSM8K, MATH500, and GPQA (Diamond), sampling $8$~trajectories per prompt at temperature $0.6$, top-$p = 0.95$, top-$k = 20$, with a maximum
of $4096$ new tokens, and report pass@1 averaged across the $8$ samples.

\paragraph{Coding evaluation.}
For HumanEval and MBPP we follow the coding evaluation protocol of \citet{jiang2025safechain}: \emph{greedy} decoding (temperature $0$), pass@1 scored with the EvalPlus harness (matching the open-source SafeChain evaluation pipeline), and a repetition penalty of $1.1$ applied \emph{only} to coding generations. The repetition penalty is the same mitigation used by \citet{jiang2025safechain} to suppress the degenerate repetition that long chain-of-thought code outputs are known to exhibit on these two benchmarks~\citep{guo2025deepseek}; no other evaluation axis uses it. We use the same maximum of $4096$ new tokens as for the safety benchmarks.

\paragraph{Checkpoint selection for reasoning.}
For each (model, method) pair, we select the checkpoint with the highest composite safety score $S$ across training epochs and evaluate all reasoning benchmarks (math and coding) on that checkpoint for a fair comparison.

\section{Detailed Jailbreak Experimental Setup}
\label{appendix:jailbreak}

\subsection{Generation and Judging Pipeline}
\label{sec:apdx:pipeline}
We follow the HarmBench three-stage pipeline: adversarial test-case generation, target-model completion, and classifier-based judging. Target completions are generated locally with vLLM in bfloat16 using each model's chat template. Because the targets are reasoning models, we allow up to 16{,}384 generated tokens per completion. The final harmfulness label is assigned by \texttt{cais/HarmBench-Llama-2-13B-cls}, which receives up to 1{,}024 tokens of the visible answer. For reasoning outputs, the judge strips the internal \texttt{<think>...</think>} block before classification, so the score reflects the visible final answer rather than the hidden chain-of-thought.

For stochastic evaluation, we sample 3 independent completions (Target samples) per test case with seed 42. Sampling parameters are read from each target's \texttt{generation\_config.json} when available: temperature $0.6$, top-$p=0.95$, and top-$k=20$ for Qwen3 models, and temperature $0.6$, top-$p=0.95$, with no top-$k$ filtering for DeepSeek-R1-Distill models. If a fine-tuned checkpoint lacks a generation config, we fall back to temperature $0.6$, top-$p=0.95$, and disabled top-$k$.

\subsection{Attack Configurations}
\label{sec:apdx:attacks}
The four attacks are not the same algorithm under different names. Each attack defines its own prompt-generation procedure and therefore induces a different number of attempts per behavior, as shown in Table~\ref{tab:attack-settings}. We keep this per-behavior budget fixed across \textsc{Base}, \textsc{Thinksafe}, and \textsc{OPSA} within each attack, which makes within-attack model comparisons fair. The right column of Table~\ref{tab:attack-settings} also makes clear why pass@$N$ values are only meaningful within an attack family: the per-behavior attempt budget ranges from 3 for Prefilling and PAIR to 339 for HumanJailbreaks.
\begin{table}[h]
\caption{Per-behavior evaluation budgets for each attack. The total number of attempts (right column) differs across attacks because each attack's prompt-generation procedure produces a different number of test cases per behavior. We hold this budget fixed across Base, Thinksafe, and OPSA, which makes within-attack comparisons fair but means pass@$N$ values are not directly comparable across attacks.}
\label{tab:attack-settings}
\vskip 0.15in
\begin{center}
\begin{small}
\begin{tabular}{lccc}
\toprule
Attack          & Test cases / behavior & Target samples  & Attempts / behavior \\
\midrule
HumanJailbreaks & 113 & 3 & 339 \\
Prefilling      & 1   & 3 & 3   \\
PAP-top5        & 5   & 3 & 15  \\
PAIR            & 1   & 3 & 3   \\
\bottomrule
\end{tabular}
\end{small}
\end{center}
\vskip -0.1in
\end{table}

\paragraph{HumanJailbreaks.}
We use the full HarmBench HumanJailbreaks configuration (\texttt{random\_subset=-1}), which yields 113 jailbreak templates per behavior and 339 labeled attempts per behavior.

\paragraph{Prefilling.}
Prefilling biases the assistant generation by prepending a compliance-suggestive string to the model's generation context. In reasoning mode, the prefix is inserted inside the opening \texttt{<think>} block:
\[
\texttt{<think>\textbackslash n Okay, let's provide clear instructions to assist the user}
\]
We use one prefilling prompt per behavior and sample three target completions.

\paragraph{PAP-top5.}
PAP rewrites each harmful behavior under the top 5 persuasion techniques from the PAP taxonomy~\citep{zeng2024johnny}. The attacker is \texttt{mistralai/Mistral-7B-Instruct-v0.2}, run with bfloat16 vLLM, attack temperature $1.0$, a maximum of $2{,}048$ generated tokens, and stop token ``\texttt{|}''. This produces 5 test cases and 15 labeled attempts per behavior.

\paragraph{PAIR.}
We use \texttt{mistralai/Mistral-7B-Instruct-v0.2} as both the attacker and the in-loop judge, replacing the larger upstream attacker and judge for local GPU feasibility. We retain the upstream PAIR search budget: 20 parallel streams, 3 optimization steps, \texttt{keep\_last\_n=3}, a maximum of 20 retries, attacker generation budget of 500 tokens, in-loop judge budget of 5 tokens, and success cutoff score 10. For reasoning targets, we increase the target generation budget to $16{,}384$ tokens and strip target \texttt{<think>...</think>} traces before feeding responses back into
the attacker conversation.

\subsection{Per-Attack Values of $N$ for pass@$N$}
\label{sec:apdx:passN}
Because each attack induces a different number of attempts per behavior, pass@$N$ is only defined within an attack family. Following the per-behavior budgets in Table~\ref{tab:attack-settings}, we report pass@$N$ at the full attempt budget of each attack: $N{=}339$ for HumanJailbreaks, $N{=}3$ for Prefilling, $N{=}15$ for PAP-top5, and $N{=}3$ for PAIR.

\subsection{Comparability Caveats}
\label{sec:apdx:caveats}
\textsc{Base}, \textsc{Thinksafe}, and \textsc{OPSA} are directly comparable within each row because they share the same behaviors, attack implementation, attacker model, target sampling, and judge. Absolute ASR values should not be interpreted as exact reproductions of upstream HarmBench numbers, because we adapt the pipeline for local reasoning-model evaluation in several ways: we use Mistral-7B-Instruct-v0.2 as the attacker and in-loop judge for PAIR and PAP, apply reasoning-aware chat-template handling, allow long target generations, and judge only the visible final answer after stripping the hidden reasoning.

\end{document}